\documentclass[journal]{IEEEtran}

\makeatletter
\def\@citex[#1]#2{\leavevmode
	\let\@citea\@empty
	\@cite{\@for\@citeb:=#2\do
		{\@citea\def\@citea{,\penalty\@m\ }%
			\edef\@citeb{\expandafter\@firstofone\@citeb\@empty}%
			\if@filesw\immediate\write\@auxout{\string\citation{\@citeb}}\fi
			\@ifundefined{b@\@citeb}{\hbox{\reset@font\bfseries ?}%
				\G@refundefinedtrue
				\@latex@warning
				{Citation `\@citeb' on page \thepage \space undefined}}%
			{\@cite@ofmt{\csname b@\@citeb\endcsname}}}}{#1}}
\makeatother

\ifCLASSINFOpdf
\else
\fi
\usepackage{array}
\newcolumntype{P}[1]{>{\centering\arraybackslash}p{#1}}
\usepackage{tabularx}
\usepackage{multirow}
\usepackage{xcolor,colortbl}
\usepackage{graphicx}
\usepackage[export]{adjustbox}
\usepackage{hhline}
\usepackage[flushleft]{threeparttable}

\usepackage{bibentry} 
\newcommand\T{\rule{0pt}{3.5ex}}       
\newcommand\B{\rule[-3.5ex]{0pt}{0pt}} 

\IEEEoverridecommandlockouts          
\makeatletter
\newcommand*\titleheader[1]{\gdef\@titleheader{#1}}
\AtBeginDocument{\let \st@red@title \@title \def \@title{\bgroup \normalfont \footnotesize \flushleft \@titleheader \par \egroup \vskip0em \st@red@title}}
\makeatother

\title{Gesture Recognition in Robotic Surgery: a Review}

\titleheader{
B. van Amsterdam, M. Clarkson and D. Stoyanov, "Gesture Recognition in Robotic Surgery: a Review," in IEEE Transactions on Biomedical Engineering, 2021. DOI: 10.1109/TBME.2021.3054828.\\
\copyright2021 IEEE. Personal use of this material is permitted. Permission from IEEE must be obtained for all other uses, in any current or future media, including reprinting/republishing this material for advertising or promotional purposes, creating new collective works, for resale or redistribution to servers or lists, or reuse of any copyrighted component of this work in other works.}

\author{
Beatrice van Amsterdam$^{1}$, Matthew J. Clarkson$^{1}$, Danail Stoyanov$^{1}$
\thanks{The work was supported by the Wellcome/EPSRC Centre for Interventional and Surgical Sciences (WEISS) [203145Z/16/Z]; the Engineering and Physical Sciences Research Council (EPSRC) [EP/P027938/1, EP/R004080/1, EP/P012841/1]; the Royal Academy of Engineering Chair in Emerging Technologies scheme; the EPSRC i4Health CDT (EP/S021930/1).}
\thanks{$^{1}$Beatrice van Amsterdam, Matthew J. Clarkson and Danail Stoyanov are with the Wellcome/EPSRC Centre for Interventional and Surgical Sciences (WEISS), University College London, UK. {\tt\small beatrice.amsterdam.18@ucl.ac.uk}}
}

\begin{document}

\maketitle

\begin{abstract}
Objective: Surgical activity recognition is a fundamental step in computer-assisted interventions. This paper reviews the state-of-the-art in methods for automatic recognition of fine-grained gestures in robotic surgery focusing on recent data-driven approaches and outlines the open questions and future research directions. 
Methods: An article search was performed on 5 bibliographic databases with combinations of the following search terms: robotic,  robot-assisted, JIGSAWS, surgery, surgical, gesture, fine-grained, surgeme, action, trajectory, segmentation, recognition, parsing. 
Selected articles were classified based on the level of supervision required for training and divided into different groups representing major frameworks for time series analysis and data modelling. 
Results: A total of 52 articles were reviewed. The research field is showing rapid expansion, with the majority of articles published in the last 4 years. Deep-learning-based temporal models with discriminative feature extraction and multi-modal data integration have demonstrated promising results on small surgical datasets. Currently, unsupervised methods perform significantly less well than the supervised approaches. 
Conclusion: The development of large and diverse open-source datasets of annotated demonstrations is essential for development and validation of robust solutions for surgical gesture recognition. While new strategies for discriminative feature extraction and knowledge transfer, or unsupervised and semi-supervised approaches, can mitigate the need for data and labels, they have not yet been demonstrated to achieve comparable performance. Important future research directions include detection and forecast of gesture-specific errors and anomalies. 
Significance: This paper is a comprehensive and structured analysis of surgical gesture recognition methods aiming to summarize the status of this rapidly evolving field.
\end{abstract}


\IEEEpeerreviewmaketitle

\section{Introduction}

Robot-assisted minimally invasive surgery (RMIS) is now well established in clinical practice as a means of enhancing surgical instrumentation and ergonomics with respect to conventional laparoscopic approaches \cite{Moorthy2004}. Surgical specialisations like gynecology and urology have driven the adoption of RMIS, but diffusion of the robotic approach is increasing as hospitals become equipped with surgical robotic systems, surgeons trained to use them and both the capital and running costs reduce \cite{Sheetz2020}. In addition to better surgical instrumentation, surgical robots can potentially be platforms for enabling data driven solutions for better OR capabilities \cite{Chadebecq2020}.

Besides digital video, surgical robots can capture quantitative instrument motion trajectories during surgical interventions enabling analysis of surgical activity that is not possible with traditional instrumentation.
These new data streams have paved the way for data-driven computational models in computer assisted interventions (CAI) and surgical data science (SDS) \cite{Maier2017} to facilitate better procedural understanding and intra-operative support, including the possibility of linking back to the robot control loop to achieve surgical automation. 
The resultant data has also allowed research into automated surgical competence and skill assessment \cite{Rosen2001, Rosen2002, Reiley2009} and adaptive surgical training \cite{Vaughan2016}, aiming at quantifying surgical process without expert monitoring.

\begin{figure}[t!]
	\centering
	\includegraphics[width=0.9\columnwidth]{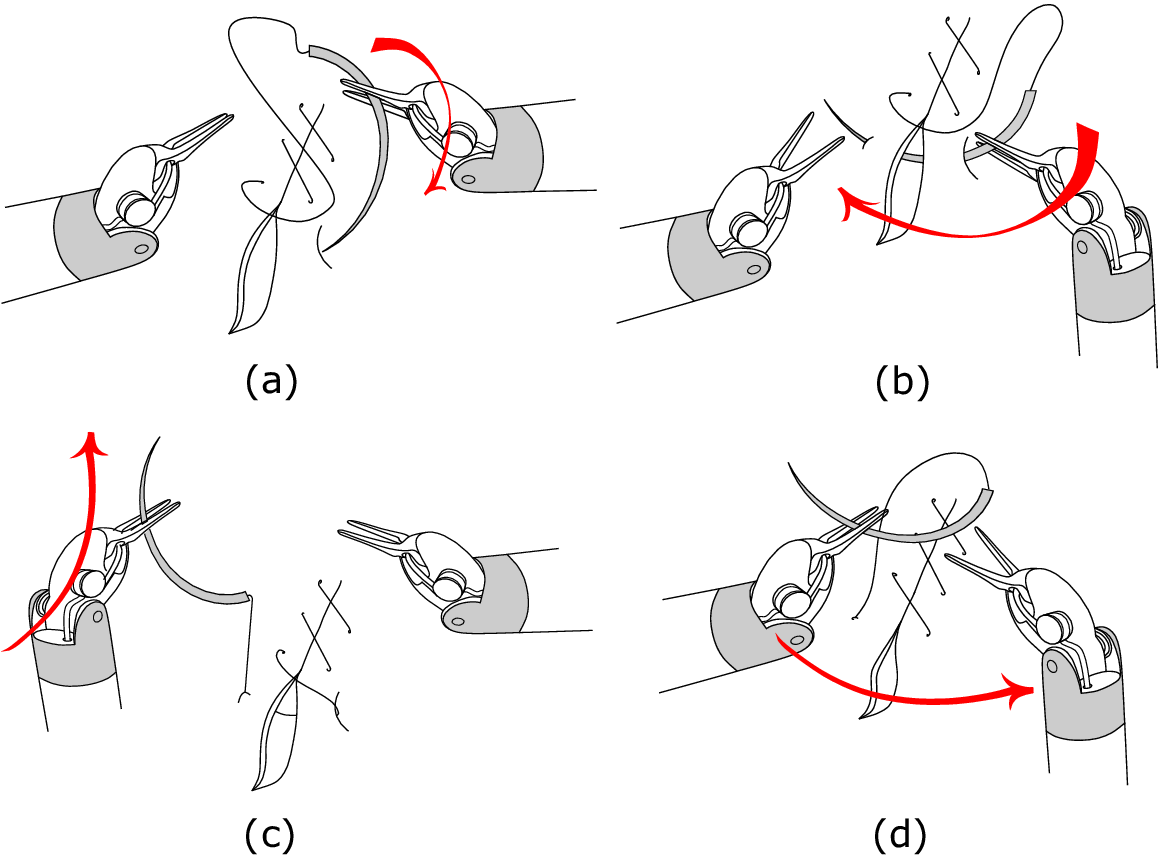}
	\caption[Surgeme examples.]{Surgeme examples for the suturing task: (a) \textit{positioning needle tip on insertion point}, (b) \textit{pushing needle through the tissue}, (c) \textit{pulling needle out of tissue}, (d) \textit{transferring needle from left hand to right hand}.}
	\label{fig:surgeme}
\end{figure}

\begin{figure}[t]
	\centering
	\includegraphics[width=\columnwidth]{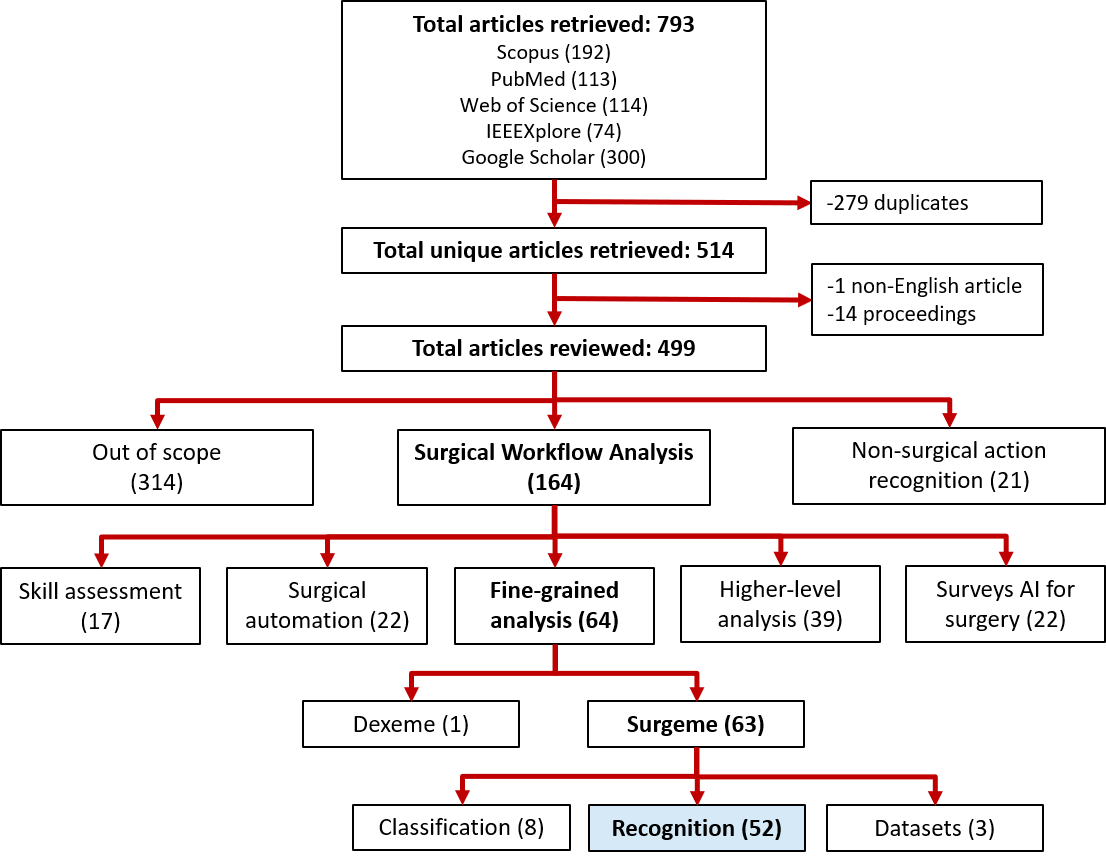}
	\caption[Selection strategy.]{Selection strategy. We identified a total of 52 articles on automatic recognition of fine-grained gestures in robot-assisted surgery, which is primarily laparoscopic and technically challenging for recognition due to environment deformations, camera motion and illumination changes.}
	\label{fig:searchSchema}
\end{figure}

Understanding surgical process has often been approached by decomposition into pre-defined procedural segments at different granularity levels. 
Starting from the finest level, surgical processes have been hierarchically decomposed into \textit{dexemes}, \textit{surgemes} (generally called \textit{gestures}, Fig. \ref{fig:surgeme}), \textit{activities}, \textit{phases}, \textit{procedures} and \textit{states} \cite{Lalys2014, Dergachyova2017}. Unfortunately, the lack of a universal taxonomy and of clear segmentation criteria has led researches to propose different decomposition strategies (e.g. semantic \cite{Vedula2016} vs event-based \cite{Novi2012} labelling) and alternative motion dictionaries \cite{Gao2014, Vedula2016, Chen2018}, resulting in conceptual inconsistencies and a lack of standardised terminology.

Segmentation at different levels highlights distinct aspects of the surgical procedure, leading to a range of applications. For example, higher-level situation-awareness (state, phase or activity recognition) is well suited for workflow optimization, scheduling and resource management \cite{Maktabi2017}. 
Recognition of fine-grained motion (dexemes, surgemes) is instead better suited for low-level analysis, assessment and reproduction of dexterous motion. While dexemes describe short motion segments devoid of medical sense (e.g. turning left) \cite{Dergachyova2017}, surgemes represent surgical gestures made with a specific purpose (e.g. grabbing the needle) and are the focus of this review. 
Kinematic trajectory segmentation into gestures has been shown to have potential to describe technical skill development \cite{Gao2014, Vedula2016, Chen2018} in a more quantitative manner than relying on global performance measures as time to completion or total path length \cite{Judkins2009}. Gesture recognition also finds application in surgical automation because short motion segments are less complex and therefore easier to learn and generalize than long surgical tasks, and can be regarded as modular blocks of motion to compose and reuse \cite{Nagy2019}.

Fine-grained analysis of surgical demonstrations is however non-trivial and presents a number of challenges. 
When generated manually, gesture annotations are costly, time consuming and subject to personal interpretation of multiple participants. Due to subjectivity and smooth transitions between gestures, boundaries between consecutive segments are often not clearly defined and it is difficult, even for a single annotator, to generate consistent segmentations as data grow over time.
For this reason significant research effort is aimed at automating the annotation process utilizing both video and instrument kinematics, either individually or in combination \cite{Tao2013, Lea2016tcnECCV, dipietro2016}.
Automation remains very challenging due to the complexity of surgical tasks and the presence of different variability factors, including surgeon's skill level, operative style, type of procedure and patient-specific anatomy, but could help to efficiently generate the large amount of training data required for robust CAI implementations.
Beyond off-line generation of training datasets for surgical automation and training, automatic recognition can be exploited for online applications where context-awareness is essential, such as systems for surgical process monitoring, error detection and intra-operative guidance and assistance of different kinds (e.g. robotic, visual, haptic) \cite{Yasar2020, DeRossi2019}. This will have a potentially high impact in robotic surgery, where demands for increased safety and effectiveness are particularly strong \cite{Bell2009}.

In this paper, we review existing methods for automatic recognition of fine-grained gestures in RMIS, illustrating how technical challenges have been addressed, what problems still need to be investigated and what could be the future research directions in the field. 
Comprehensive reviews on higher-level surgical workflow recognition have recently been reported and are outside the scope of this study \cite{Padoy2019, Dergachyova2017, Lalys2014}.

\begin{figure}[t]
	\centering
	\includegraphics[width=\columnwidth]{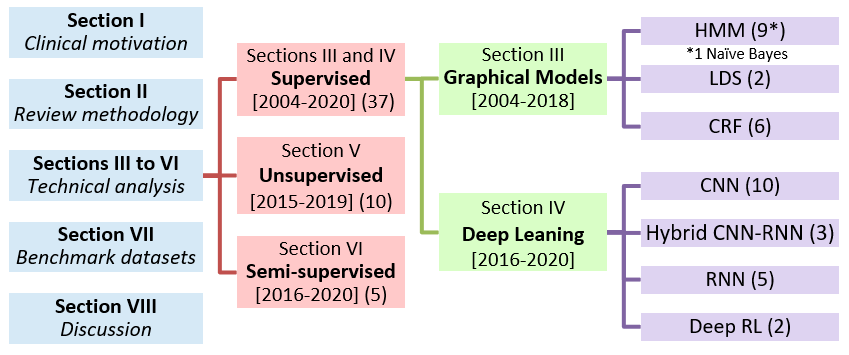}
	\caption[Review structure.]{Review structure. The selected methods were classified based on the learning paradigm (red boxes) and recognition technique (green and purple boxes).}
	\label{fig:reviewSchema}
\end{figure}

\begin{figure*}[t]
	\centering
	\includegraphics[width=\textwidth]{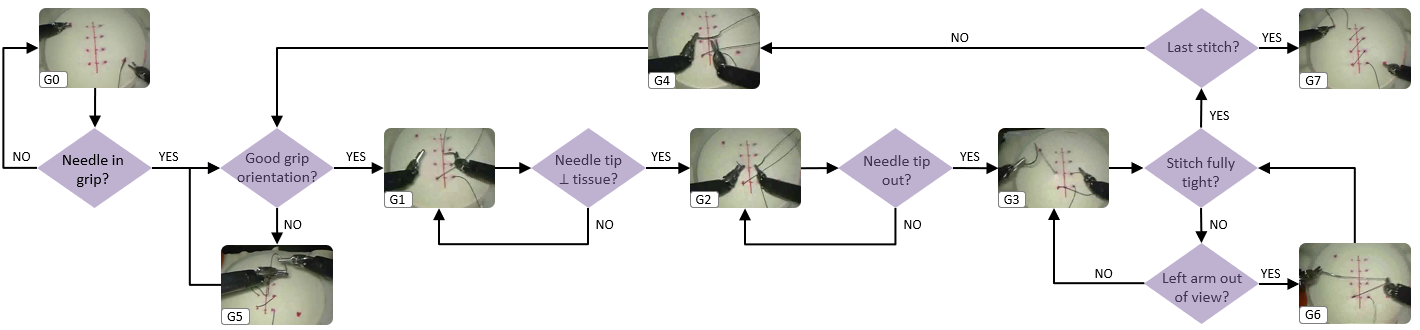}
	\caption[Suturing grammar graph.]{Example of grammar graph for a suturing demonstration performed with the right hand. 
	Gesture labels: 
	G0 = reaching needle with right hand (START),
	G1 = positioning needle with right hand,
	G2 = pushing needle in tissue with right hand,
    G3 = pulling suture with left hand,
    G4 = transferring needle from left hand to right hand,
    G5 = orienting needle on right hand,
    G6 = tightening suture with right hand,
    G7 = dropping suture (END).
    The transition between gestures depends on specific conditions, here represented in purple.
    Images from JIGSAWS \cite{Ahmidi2017}.}
	\label{fig:grammar}
\end{figure*}

\section{Methodology and structure} \label{Structure}

We used Google Scholar, Scopus, PubMed, Web of Science and IEEEXplore for searching the literature titles, abstracts and keywords with the query: ``(robotic  OR  robot-assisted  OR  JIGSAWS)  AND  (surgery OR surgical)  AND  (gesture OR fine-grained OR surgeme OR action OR trajectory)  AND  (segmentation OR recognition OR parsing) ", where JIGSAWS represents the first open source dataset for surgical gesture recognition (see Section \ref{The JIGSAWS dataset}). As Google Scholar only allows full-text searches, the 300 most relevant publications were extracted from the 47.100 retrieved items. Integrating the 5 databases and discarding all duplicates, non-English texts and full conference proceedings, our query delivered a total of 499 matches in October 2020.

We adopt the term ``recognition" to refer to the joint problem of ``segmentation" and ``classification" of surgical gestures, i.e. the simultaneous segmentation of surgical demonstration into distinct motion units and the classification of these units into meaningful action categories. 
While computer-vision methods for classifying videos with known action boundaries have been successful, temporal segmentation of multiple action instances is still a hard-won problem. Classification of trimmed videos also finds limited applicability in surgical scenarios due to gesture annotation costs, thus research has been fairly limited (only 8 studies were retrieved with our query) and is not presented in this review.  
We also excluded articles focused on coarser (``higher-level") granularity and studies that use non-novel recognition methods for skill assessment or surgical automation.  
After a selection process as illustrated in Fig. \ref{fig:searchSchema}, we identified a total of 52 publications to be reviewed. 

We use the field decomposition shown in Fig. \ref{fig:reviewSchema} to organise our analysis. Supervised approaches are classified into two groups representing major frameworks for time series analysis and data modelling: probabilistic graphical models (Section \ref{Probabilistic graphical models}) and deep learning (Section \ref{Deep Learning}). We then describe efforts to mitigate the need for manual annotations (Sections \ref{Unsupervised learning} and \ref{Semi-supervised Learning}). 
A subdivision of the methods by data modality is instead highlighted in Tables \ref{tab1}, \ref{tab2}, \ref{tab3} and \ref{tab4} of Section \ref{Method comparison}.

\section{Supervised learning - graphical models} \label{Probabilistic graphical models}

Inspired by time series analysis for speech recognition, structured temporal models such as probabilistic graphical models of temporal processes have been extensively used for gesture recognition in robotic surgery \cite{Ahmidi2017}. Similarly to the natural language, where syntactic rules govern the generation of words and sentences, the surgical language can be decomposed into surgical motion units at different granularity levels, and probabilistic grammars can be identified to describe the structure of higher-level surgical tasks (Fig. \ref{fig:grammar}).

There is often significant flexibility in choosing the structure and parameterization of a graphical model. Given a sequence of observations and corresponding output labels, generative models such as Hidden Markov Models (HMM) and Linear Dynamical Systems (LDS) explicitly attempt to capture the joint probability distribution over inputs and output. Discriminative models like Conditional Random Fields (CRF), on the other hand, attempt to model the conditional distribution of the output labels given the input data. The advantage with respect to generative models is that dependencies in the input features don’t need to be modelled, so that conditional models can have simpler structure \cite{Sutton2011}.

\begin{table*}[ht] 
    \caption{Probabilistic graphical models for surgeme recognition.} \label{tab:GMsummary}
    \begin{tabular}{m{1.5cm}m{3.5cm}m{5.5cm}m{5.5cm}} 
        \hline 
        \multicolumn{2}{l}{\textbf{Feature extraction}} & \textbf{Pros} & \textbf{Cons} \\
        \hline
        \multirow{3}{1em}{\T \includegraphics[width=0.58in]{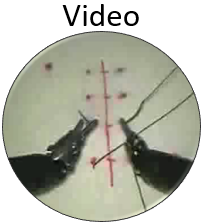}}  &  \T Bag-of-features \cite{Tao2013} & \T Spatio-temporal information & \T Hand-crafted  \\
		& \T Semantic features \cite{Lea2015semantic, Lea2016convolutional} &  \T Semantic information & \T Hand-crafted, localized information  \\
		& \T CNN \cite{Lea2016, Rupprecht2016} \B & \T Spatio-temporal information, data-derived \B & \T Short attention span \B \\
		\hline
		\multirow{3}{1em}{\includegraphics[width=0.58in]{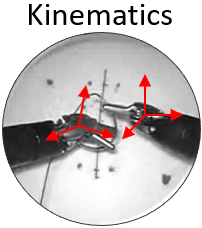}} &  \T LDA \cite{Murphy2004, Lin2006, Reiley2008, Varadarajan2009} &  \T Enhanced (sub-)gesture separation & \T No temporal information  \\
		& \T Sparse Codings \cite{Tao2012, Sefati2015, Mavroudi2018} &  \T Expressive and compact data representation & \T No temporal information  \\
		& \T Convolutional filters \cite{Lea2016convolutional} \B &  \T Sub-gesture analysis, temporal information \B & \T Short attention span \B \\
		\hline
		\multicolumn{2}{l}{\textbf{Recognition model}} & \textbf{Pros} & \textbf{Cons} \\
        \hline
        \multirow{3}{1em}{\includegraphics[width=0.6in]{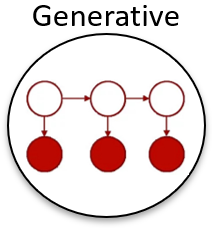}} &  \T Na\"ive Bayes \cite{Lin2006} &  \T Simple model & \T Frame-wise classification  \\
		& \T HMM \cite{Murphy2004, Reiley2008, Varadarajan2009, Tao2012, Sefati2015, Selvaggio2018} &  \T Transition model between (sub-)gestures & \T Markov assumption, short attention span  \\
		& \T LDS \cite{Varadarajan2011, Varadarajan2011VAR} &  \T Continuous hidden state & \T Local linearity hypothesis, short attention span \B  \\
		\hline
		\hspace{1cm}
		\includegraphics[width=0.65in]{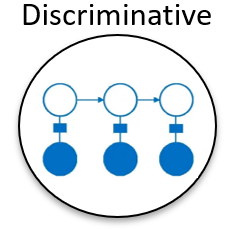} & CRF \cite{Tao2013, Lea2015semantic, Mavroudi2018, Lea2016, Lea2016convolutional, Rupprecht2016} &  Segment-to-segment dependencies (MsM-CRF, SC-CRF) & No temporal regularization of adjacent samples (SC-CRF), lack of longer-range dependencies \\
		\hline
    \end{tabular}
\end{table*}

\subsection{Hidden Markov Models}

HMMs were first used to classify motion segments obtained by thresholding the kinematic signals, which had limited recognition rate but could be significantly improved with appropriate data transformation including normalization to zero-mean and unit-variance, and Linear Discriminant Analysis (LDA) \cite{Murphy2004}.
Normalization compensates for different units of measure in the kinematic features, while LDA reduces the data dimensionality and model complexity, which is important for robust parameter estimation when only little data is available. LDA's ability to enhance the separation between classes is another desirable property for fine-grained gesture recognition, where different classes exhibit similar motion and visual appearance.

Highly discriminative feature extraction is key to the recognition problem, so that even simple frame-wise classifiers, such as the Na\"ive Bayes classifier, can provide high recognition accuracy \cite{Lin2006}. 
Following \cite{Murphy2004}, LDA was successfully applied to the high-dimensional kinematic signals recorded from the da Vinci surgical robot in combination with a continuous HMM for gesture recognition, composed of basic HMMs representing different surgemes and trained individually \cite{Reiley2008, Grigoli2018}.
Recognition rate of such models further improves promoting discrimination between sub-gestures (dexeme) rather than entire gestures (surgeme), as this helps to capture the evolution and internal variability of each segment \cite{Varadarajan2009}. Automatic derivation of the optimal HMM topology (i.e. the optimal number of hidden states) can also help to discover skill-related and context-specific sub-gestures. While yielding accurate recognition on new trials of seen users, data-derived HMMs may also introduce user-specific sub-gestures and therefore overfit on observed subjects \cite{Varadarajan2009}.

Alternative data representation and reduction techniques investigated for recognition include Factor-Analyzed Hidden Markov Models (FA-HMM), which yield better recognition compared to standard HMMs with LDA \cite{Varadarajan2011}. Sparse Hidden Markov Models (SHMM) were also applied to the recognition of surgical gestures \cite{Tao2012, Sefati2015}. Available implementations of SHMM represent each surgeme with a different hidden state, whose observations are modelled as sparse linear combinations of atomic motions. The advantage with respect to conventional discrete \cite{Reiley2009} or Gaussian \cite{Rosen2002, Lin2006, Varadarajan2009} observation models relies in the sparsity property, which guarantees more robust learning with high-dimensional signals such as the robot kinematics. 
While each surgical gesture can be associated to a different dictionary that is learnt independently \cite{Tao2012}, more compact solutions propose unique motion dictionaries shared among all gestures \cite{Sefati2015}.

Finally, interaction forces measured between the robotic tool tip and the environment have been employed as complementary data modality to the robot kinematics \cite{Selvaggio2018}. Force information allows one to assign physical interpretation to HMM state observations. 

\subsection{Linear Dynamical Systems}

A fundamental drawback of the HMM is its reliance on discrete switching states to model temporal dependencies in the input signals, thus failing to capture more complex correlations in the continuous kinematic and video data. Switched Linear Dynamical Systems (S-LDS) were then proposed to model temporal dependencies between adjacent observations \cite{Varadarajan2011}. S-LDSs approximate non-linear systems, such as demonstrations of complex surgical tasks, using a suitable number of linear systems. The data distribution within each gesture is thus modelled through the linear evolution of a continuous hidden state, while the transition between different regimes relies on a discrete switching hidden state. Thanks to improved dynamics modelling, recognition with S-LDS outperformed FA-HMM significantly \cite{Varadarajan2011}.

S-LDS models can be further extended to capture the dynamics of past observations samples, thus becoming Switched Vector Auto-Regressive models (S-VAR). By imposing specific structure on the model parameters, with certain parameters depending on the discrete latent state and others which are time-invariant, S-VAR models can learn both gesture-dependent and global dynamics in the kinematic signals, resulting in improved recognition accuracy \cite{Varadarajan2011VAR}.

\begin{figure*}[t!]
	\centering
	\includegraphics[width=\textwidth]{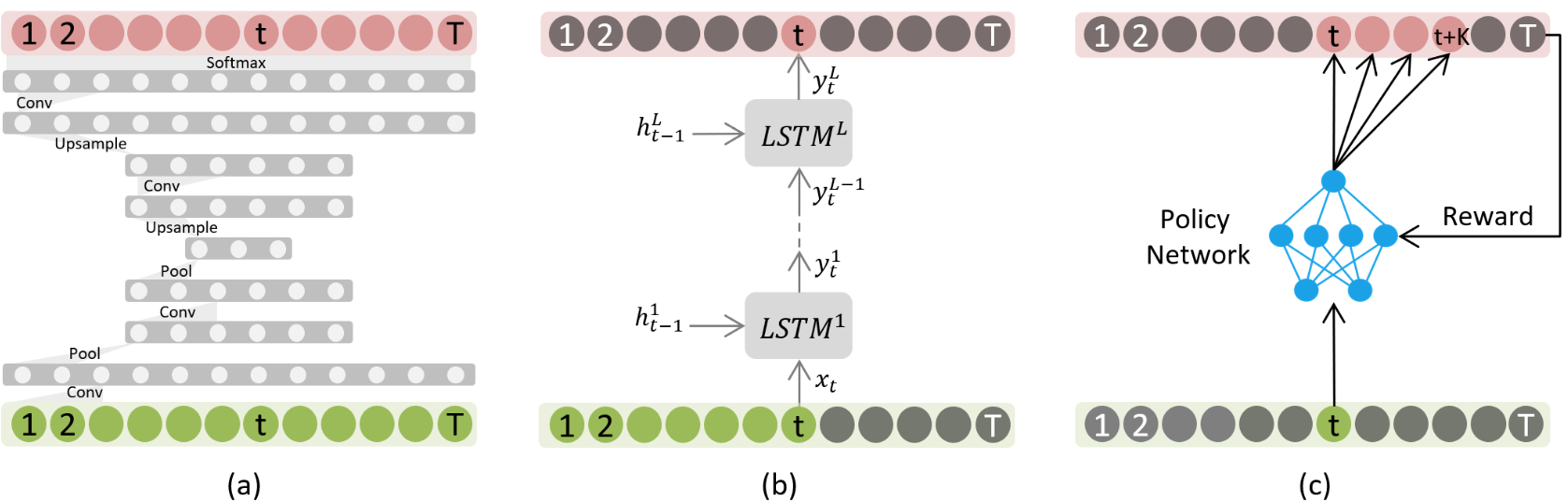}
	\caption[Deep Learning]{Deep learning for surgeme recognition: (a) temporal convolutional, (b) recurrent and (c) reinforcement learning strategies.
    Input kinematic or video sequences ($t=1:T, T=\ $sequence length) are represented at the bottom and corresponding predictions at the top. Highlighted in green are the input samples that influence the current system predictions (red circles).
	a: TCN schematic (figure reproduced from \cite{Lea2017tcnCVPR}). All input samples are processed simultaneously through layers of temporal convolutional filters to deliver predictions over the entire sequence. b: Generic multi-layer RNN ($l=1:L, L=\ $number of layers). The input samples ($x_t$) are processed sequentially in time, with each prediction ($y_t^L$) depending in principle on all the previous samples through memory of previous hidden states ($h_{t-1}^l$). c: Gesture recognition with reinforcement learning \cite{Liu2018}. At each time stamp, predictions are generated over K frames ahead.}
	\label{fig:TCN-LSTM-RL}
\end{figure*}

\subsection{Conditional Random Fields}

CRFs were first used to recognize surgical gestures in combination with video data. Inspired by related work on video-based surgeme classification \cite{BejarHaro2012}, bag of spatio-temporal visual features extracted around space-time interest points or dense trajectories were combined with the kinematic signals in input to a Markov/semi-Markov recognition model (MsM-CRF), which is able to capture frame-level and segment-level dependencies in the input signals \cite{Tao2013}.

Long-range temporal dependencies were also modelled with the Skip-Chain conditional random field (SC-CRF) \cite{Lea2015semantic}, incorporating a new skip-length data potential into the CRF energy function to capture sample dependencies at d frames of distance, where d is a skip-length hyper-parameter. 
In addition, a set of semantic features were extracted from the laparoscopic videos based on the distance of each surgical tool to the closest object in the surgical environment. The authors argued that in constrained environments such as surgical training stations, where the set of objects is known, semantic features encoding object-based information could help to discriminate surgical gestures better than abstract visual features as in \cite{Tao2013}.

As for HMMs, efforts were spent on improving data representation and feature extraction for CRFs. 
Inspired by \cite{Sefati2015}, kinematic signals were modelled as sparse linear combinations of atomic motions from an over-complete motion dictionary shared among all gestures \cite{Mavroudi2018}. The motion dictionary was learnt jointly with a SC-CRF to capture temporal dependencies and generate more discriminative signal representations, which is valuable to fine-grained analysis.
Similarly to \cite{Varadarajan2009}, surgical kinematics were also analysed at the sub-gesture level. Temporal convolutional filters were used to model the evolution of ``latent action primitives" composing each surgical gesture and capture variability within surgemes \cite{Lea2016convolutional}.
The real kinematic data were then replaced with virtual kinematic signals estimated from the video data through a spatial CNN, addressing the situation where kinematic information is available during training, but only video data are available for testing \cite{Rupprecht2016}. 
Convolutional filters were later employed in a similar but more powerful multi-stage framework based on deep convolutional neural networks for spatio-temporal feature extraction from the raw video data (ST-CNN) \cite{Lea2016}. 
Nevertheless, feature extraction strategies as in \cite{Lea2016convolutional} and \cite{Lea2016} are only able to capture local temporal correlations in the observations. Modelling of long-range dependencies is entirely entrusted to the high-level temporal classifier (SC-CRF and sM-CRF respectively). A summary of the presented graphical models for surgeme recognition is reported in Table \ref{tab:GMsummary}.

\section{Supervised learning - deep learning} \label{Deep Learning}

Recently, Deep Neural Networks (DNN) have emerged as the go-to choice for powerful feature extractors that can be learnt automatically from the data, in contrast with hand-crafted filters based on domain-specific prior knowledge \cite{deepVStrad2019}. In DNNs, layers of computational units are stacked to generate features at different semantic levels and with increased spatio-temporal receptive fields. This is advantageous for action recognition because long-range spatial and temporal dependencies encoded in the input data can be captured already in early processing stages, easing the subsequent recognition task.

\subsection{Convolutional Neural Networks}

Temporal Convolutional Network (TCN) is one of the first deep convolutional models proposed for fine-grained surgeme recognition \cite{Lea2016tcnECCV}.
TCN uses a hierarchy of temporal convolutions and pooling layers to capture long-range temporal dependencies in the input data (Fig. \ref{fig:TCN-LSTM-RL}a), thus learning action dependencies in surgical demonstrations better than competing methods and producing well structured predictions.
Integration of bilinear pooling with learnable weights further boosts recognition performance incorporating second-order statistics adaptable to the data \cite{Zhang2019}.
As opposed to RNNs, TCN predictions are computed simultaneously for each time stamp and training is much faster.

Built on a similar encoder-decoder backbone, more recent solutions rely instead on two-stream processing: one stream to capture multiscale contextual information and action dependencies in the input sequence, the other to extract low-level information for precise action boundary identification \cite{Lei2018, Wang2019}. 
Basic temporal convolutions were replaced with a set of Deformable Temporal Residual Modules (TDRM) to capture the temporal variability of human actions and merge the two streams at multiple processing levels \cite{Lei2018}.
Alternatively, atrous (aka dilated) temporal convolutions and pyramid pooling were used to explicitly generate multi-scale encoding of the input data, that were merged with the low-level local features just before the decoding phase \cite{Wang2019}. 
Both architectures were able to improve consistently both frame-wise and segmental evaluation scores (see Section \ref{Method comparison} for a description and explanation of the most common evaluation metrics) upon the competing methods.

While pooling operations help to increase the temporal receptive field of a network, they are also responsible for partial loss of fine-grained information and less precise identification of the gesture boundaries. Stacking multiple layers of dilated convolution with increasing dilation factor was proposed as alternative strategy to model long range temporal dependencies between surgical gestures \cite{Menegozzo2019}. Gesture predictions can be further refined in a multi-task, multi-stage framework where each stack of dilated convolutions is applied to the output of the previous stage, and the whole system is trained on the sum of all stage losses, as well as on the auxiliary task of surgical skill score prediction \cite{Wang2020}. Alternatively, two convolutional stages can be bridged with a self-attention kernel to encode local and global temporal dependencies in the input signals, returning highly accurate predictions with little training cost \cite{Zhang2020}.

All aforementioned architectures represent high-level temporal models that can process efficiently only low-dimensional signals such as kinematic data or pre-encoded visual features. 
The latter are often derived from the original video frames through spatial encoding techniques \cite{Lea2016}, albeit unable to capture the dynamics in surgical demonstrations.
To address this issue, spatio-temporal feature extraction strategies from short video snippets were proposed using 3D CNNs \cite{Funke2019} or Spatial Temporal Graph Convolutional Networks (ST-GCN) \cite{Sarikaya2020}. While 3D CNNs operate directly on the video data, achieving superior recognition performance to related spatial \cite{Lea2016} and spatio-temporal \cite{Lea2016} models, ST-GCN are applied on spatio-temporal skeleton representations of the surgical tools, which are automatically estimated from the video clips. Skeleton-based features are independent of the background and thus robust to scene and context variations, making them suitable for recognition across different surgical tasks or datasets. 
Clip-based methods like 3D CNNs and ST-CNNs have however limited span of attention, which prevents them from learning long-range temporal dependencies, yielding suboptimal segmental scores.  
Integration with higher-level convolutional or recurrent models could further boost segmentation performance.

Finally, efforts were spent on improving recognition robustness to inter-surgeon data variability, associated with heterogeneity factors such as surgical style and expertise level, which could hinder generalization performance on unobserved subjects \cite{Ahmidi2017} (see Sections \ref{Evaluation metrics} and \ref{Discriminative feature extraction} for more details).
Robustness to the subject's identity was obtained via adaptive feature normalization \cite{Kaku2020}, where the feature statistics are estimated over data sub-sets recorded from the same surgeon aiming to reduce inter-surgeon data variability and ease the recognition process.

\subsection{Recurrent Neural Networks}

Recurrent Neural Networks (RNN) have been used to capture long-term non-linear dynamics in surgical kinematic data \cite{dipietro2016}. 
In-depth analysis and comparison of different architectures (simple Bidirectional RNNs, Bidirectional LSTMs, Bidirectional GRUs and Bidirectional Mixed History RNNs) showed that LSTMs and GRUs, which were conceived to alleviate the vanishing gradient problem and can be trained more robustly, are less sensitive to hyper-parameter choice and achieve the best recognition performance \cite{DiPietro2019}.

Since predictions are computed sequentially in time, RNNs can naturally handle input signals of different duration and can operate in real time.
They also maintain memory mechanisms so that predictions depend in principle on all the previous time stamps (Fig. \ref{fig:TCN-LSTM-RL}b). Their temporal span of attention, however, is in practice much shorter and they fail to capture multi-scale temporal dependencies in the observations \cite{Gurcan2019}.
To address this issue, Multi-Scale Recurrent Neural Network (MS-RNN) \cite{Gurcan2019} extracts multi-scale temporal information from the kinematic signals by means of hierarchical convolutions with wavelets, which are then given as additional inputs to a LSTM for gesture recognition. MS-RNN is trained end-to-end to generate compact descriptors and lessen LSTM susceptibility to overfitting.

Alternative approaches have been recently proposed to regularize training of recurrent architectures and avoid overfitting, such as using data augmentation to improve the network robustness to rotation \cite{Itzkovich2019} or multi-task learning to jointly recognize surgical gestures and estimate the progress of the surgical task, in order to introduce ordering relationships and temporal context explicitly into the recognition process \cite{VanAmsterdam2020}.

\subsection{Hybrid Convolutional-Recurrent Networks}

Temporal Convolutional and Recurrent Network (TricorNet) \cite{Ding2017} represents the first hybrid network combining convolutional and recurrent paradigms to recognise surgical gestures. TricorNet features a temporal convolutional encoder similar to \cite{Lea2016tcnECCV} to learn local motion evolution, followed by a hierarchy of bidirectional LSTMs to capture long-range temporal dependencies and decode the prediction sequence.
Since recurrent cells are in principle able to carry long-term memory of the past observations, the authors argued that hybrid solutions could better model high-level action dependencies than purely convolutional systems with fixed-size local receptive field.

Rather than combining different models sequentially, Fusion-KV \cite{Qin2020} performes multi-modal learning with multiple recognition models (CNN-TCN, TCN, LSTM, Random Forest, Support Vector Machine) operating in parallel on three different data sources: kinematics, video and system events collected from the robotic platform (e.g. camera follow, instrument follow, surgeon head in/out of the console, etc). Predictions from all streams are fused through a weighted voting scheme that takes into account recognition performance of each individual stream on the training data. The strengths of different methods and data are thus combined to deliver improved recognition performance. 
Fusion-KV was later incorporated in a mult-task sequence-to-sequence framework for surgical gesture and instrument trajectory prediction, where gesture labels and instrument positions are estimated over multiple steps in the future given a past observation window \cite{Qin2020b}. Compact features are first extracted from multiple data sources (full video frames, region-of-interest around surgical tools and kinematic data) and concatenated into a single multi-modal feature vector. Multi-modal features for an observation window in the past and corresponding gesture labels estimated with Fusion-KV are then given as input to an attention-based sequence-to-sequence decoder for gesture prediction, showing 10-step prediction rates comparable to state-of-the-art recognition (0-step prediction) models.

\subsection{Deep Reinforcement Learning} \label{Reinforcement Learning}

Surgeme recognition can alternatively be modelled as a sequential decision-making process that can be learnt with Reinforcement Learning (RL), as recently proposed by \cite{Liu2018}.
In their work, an intelligent agent observes each surgical sequence from the beginning. At each time stamp, it selects an appropriate step size and gesture label according to a specific policy, and then moves step-size frames ahead, classifying all the frames in-between as the selected gesture (Fig. \ref{fig:TCN-LSTM-RL}c). 
The reward function for policy learning is driven by both classification error and chosen step size, introducing a bias towards larger steps to promote prediction smoothness.
While the system was able to achieve competitive results with respect to several state-of-the-art techniques, it sometimes delivered uncertain predictions, especially for rare gestures or at the segmentation boundaries. 
A tree-search component was therefore introduced to refine the low-confidence predictions from the policy network by looking into the future, improving recognition of the action boundaries \cite{Gao2020}.

\section{Unsupervised learning} \label{Unsupervised learning}

Because of the lack of labelled examples, unsupervised approaches are mostly focused on the segmentation sub-problem, i.e. the identification of motion unit boundaries within surgical demonstrations.

Transition State Clustering (TSC) \cite{Krishnan2015}, a multilevel generative model based on hierarchical Gaussian Mixture Model (GMM), was applied to surgical kinematic data to identify segmentation points pruning spurious segments generated by inconsistent motion, 
under the assumption that action order is partially consistent across surgical demonstrations.
The kinematic signals were then integrated with visual features extracted with a pre-trained CNN, improving segmentation performance \cite{Murali2016}. 
TCN methodology was later extended using Dense Convolutional Encoder-Decoder Netowork (DCED-Net) for unsupervised feature extraction from surgical videos \cite{Zhao2018}. Trained on the task of image reconstruction, DCED-Net extracts discriminative visual features using specific convolutional blocks, called Dense Blocks, to reduce information loss during dimensionality reduction.
Segmentation results were further improved with a majority vote strategy to eliminate spurious transition points exploiting both kinematic and visual information, and an iterative algorithm to merge similar adjacent segments \cite{Zhao2018}.

More complex generative models were later proposed to describe the joint probability distribution of observed kinematic data and latent action sequences.  
Surgical demonstrations were for instance modelled as Markov decision processes learnt through imitation learning with options \cite{Fox2017}, where complex policies are broken down into simpler low-level primitives called options. 
The model parameters were estimated with a policy-gradient algorithm aimed to maximize the likelihood of the available data and simultaneously segment them into different motion primitives.
Alternative solutions such PRISM (PRocedure Identification with a Segmental Mixture model) \cite{Goel2019} were developed under the assumption that all demonstrations are described by a common latent sequence of actions. While generalizable to non-Markov processes, PRISM performance degrades significantly with increased action ordering variability.

Other clustering techniques group the data samples based on feature similarities and temporal constraints.
Spectral clustering was tailored to temporal segmentation through specific regularization terms considering local dependencies in the motion data, where consecutive samples are likely to belong to the same action class \cite{Clopton2017}.
Good results were obtained with bottom-up clustering, where the sequential nature of motion signals can be exploited by merging neighbouring segments according to various merging and stopping criteria \cite{Despinoy2016, Fard2017}. 
In \cite{Despinoy2016}, surgical demonstrations were first segmented into a set of very fine-grained motion primitives (dexeme) based on the persistence and dissimilarity of consecutive segments between critical points in the kinematic trajectories.
A descriptive signature was then created to represent each dexeme and used for classification into surgeme categories. 
Starting instead from the finest possible segmentation of the kinematic signals \cite{Fard2017},
neighboring segments were merged iteratively using multiple compaction scores based on different distance measures (similarity between PCA results, distance between segment centres, dynamic time warping distance) and fuzzy membership scores to model gradual transitions between surgical gestures.

Rather than grouping similar data samples or segments together, opposite approaches search for optimal segmentation criteria.
Spatio-temporal and variance properties of the kinematic data can be exploited
to identify candidate segmentation points around the peaks of distance and variance profiles computed from the original signals \cite{Tsai2019unsup}.  
The method's disadvantage, however, is in the tedious tuning of threshold values.

Finally, zero-shot learning (ZSL) approaches have also shown potential for surgical gesture recognition \cite{Jones2019}. 
ZSL consists in learning a set of high-level semantic attributes describing instances of specific action classes and exploiting them to recognize instances of new classes, provided with their attribute description (signature) but no labelled examples.
\cite{Jones2019} modelled surgical gestures with dynamic signatures represented by a set of attributes that change in time according to action-specific rules.
Given a surgical demonstration, the presence or absence of those attributes were detected on each video frame and compared to the set of signatures for action matching, achieving better results than fully supervised unstructured baseline models.

\section{Semi-supervised learning} \label{Semi-supervised Learning}

While unsupervised approaches mitigate significantly labelling costs and issues, as labels are only necessary for model testing, they still show a significant gap in performance with respect to the supervised methods. 
A midway solution is represented by semi-supervised learning \cite{Gao2016align, vanAmsterdam2019, DiPietro2019one, Tsai2019}, where a small pool of annotated demonstrations is used for model training. Such demonstrations can be used to transfer gesture labels to a large set of unlabelled observations previously aligned with DTW \cite{Gao2016align}, or to initialize clustering of unlabelled data and avoid tedious parameter tuning \cite{vanAmsterdam2019}. These systems are however highly sensitive to large data variability, showing rapid drop in recognition accuracy with increased heterogeneity in terms of gesture ordering or surgical skill level respectively.

Promising results were obtained with RNNs within a two-phase pipeline comprising unsupervised representation learning and semi-supervised recognition \cite{DiPietro2019one}.
An RNN-based generative model was first trained to learn the full joint distribution over the kinematic signals, an auxiliary task that needs no additional annotations but can extract informative features from the input data. 
These features were then used by a bidirectional LSTM for gesture recognition, which was only trained on a very small number of labelled demonstrations.
Decoupling between representation learning and gesture recognition was later solved via semi-supervised interleaved training \cite{Tanwani2020}.
Meaningful visual embeddings were learnt using triplet loss to pull together images in the same action segment and push away samples from different segments, where action pseudo-labels for unlabeled training sequences were iteratively predicted by the current estimate of a recurrent network trained via cross-entropy loss. 
Thanks to robust representation learning, the gap in performance with respect to equivalent fully supervised systems was limited.

Semi-supervised learning was also explored in combination with transfer learning between surgical tasks \cite{Tsai2019}. One-dimensional projections of the kinematic signals derived from their Self-Similarity Matrix (SSM), which preserves motion patters in the data, were used to learn segmentation policies of specific surgical tasks (e.g. knot-tying) from annotated demonstrations of other tasks (e.g. suturing). In contrast to \cite{Gao2016align, vanAmsterdam2019, DiPietro2019one}, the class of the generated segments was not predicted.

\begin{figure}[t]
	\centering
	\includegraphics[width=\columnwidth]{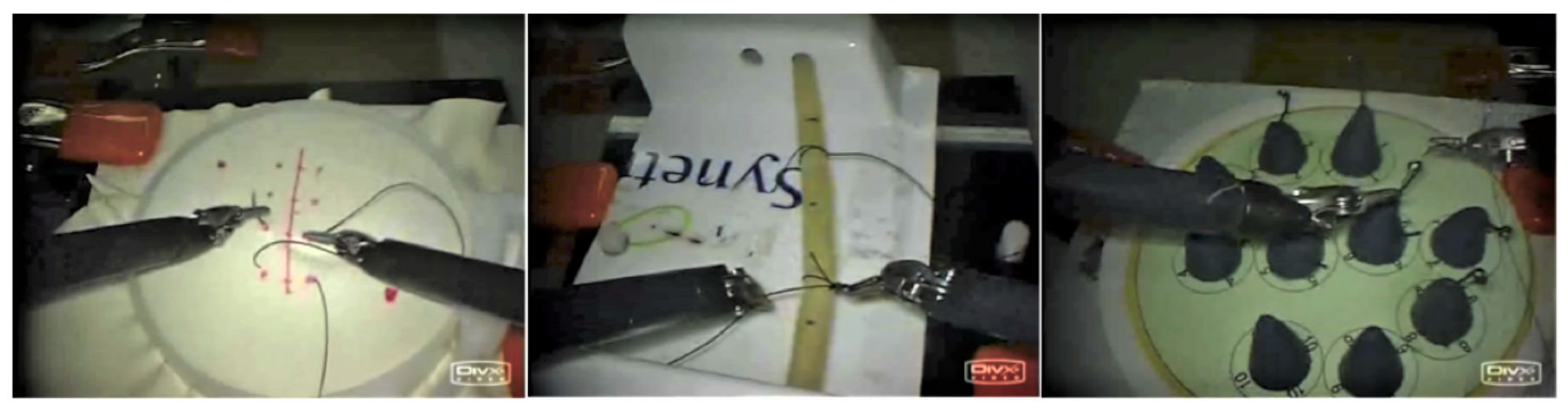}
	\caption[JIGSAWS video snapshots.]{Snapshots from suturing, knot-tying and needle passing demonstrations in JIGSAWS \cite{Ahmidi2017}.}
	\label{fig:jigsaws}
\end{figure}

\section{Benchmark datasets and evaluation protocols} \label{Method comparison}
Early techniques for surgical gesture segmentation and skill assessment were validated on study-specific data using different evaluation metrics, hindering comparative research and progress tracking in the field.
To address this issue, the JHU-ISI Gesture and Skill Assessment Working Set (JIGSAWS) \cite{Gao2014, Ahmidi2017} was released as first public dataset containing video and kinematic recordings of robotic surgical demonstrations, as well as gesture and skill annotations. 

\subsection{The JIGSAWS dataset} \label{The JIGSAWS dataset}
JIGSAWS contains synchronized video and kinematic data captured from the da Vinci Surgical System (dVSS) during the execution of elementary surgical tasks, including 39 suturing, 36 knot-tying and 20 needle-passing demonstrations (Fig. \ref{fig:jigsaws}). All trials are performed on a training phantom by eight surgeons with different robotic surgical experience: experts reported more than 100 hours of training, intermediate between 10 and 100, and novice less than 10 \cite{Gao2014}. 

\textit{Video data}: Stereo videos from the da Vinci endoscopic cameras are recorded at 30Hz and at a resolution of 640 x 480. 
Demonstrations are all observed from the same point of view, as camera movement and zoom are completely absent.
Calibration parameters are not provided.

\textit{Kinematic data}: Kinematic trajectories of the two master tool manipulators (MTMs) and patient side manipulators (PSMs) are recorded in sync with the video data.
The pose of each arm is represented by a local reference frame linked to the end-effector, and their motion is described by 19 kinematic variables that include Cartesian positions, a rotation matrix, linear velocities, angular velocities and a gripper angle. Kinematic features of different manipulators are all referred to a common coordinate system.

\textit{Annotations}: 
A common dictionary comprised of 15 action classes is used to describe all three surgical tasks in the dataset. Ground truth annotations for each trial include the sequence of gestures, boundary timestamps and the demonstrator's expertise level. 

\begin{table*}[t]
    
	\caption[Graphical Models.]{Average cross-validation scores (Accuracy and Edit*) on the suturing demonstrations of JIGSAWS - Graphical Models. \\green = kinematics, yellow = video, red = kinematics and video.}\label{tab1}
	\centering
	\footnotesize
	\renewcommand{\arraystretch}{1.5}
	\renewcommand{\arrayrulewidth}{1pt}
	
    \begin{threeparttable}
    	\begin{tabular}{m{2.9cm}|P{0.6cm}|cc|cc|m{9.0cm}}
    		\hhline{*{7}{-}}
    		
            \rowcolor{white}
    		\hspace{0.9cm} \textbf{Method} & \textbf{Year}
    		& \multicolumn{2}{c|}{\bfseries LOSO } & \multicolumn{2}{c|}{\bfseries LOUO } &
            \hspace{4.1cm} \bfseries Data\\
		
    		\rowcolor{white}
    		 &  & \textbf{Acc} & \textbf{Edit*} & \textbf{Acc} & \textbf{Edit*} &  \\
    		\hhline{*{7}{-}}
    		
    		\textbf{\textit{Composite-HMM}} \scriptsize \cite{Varadarajan2009, Ahmidi2017} & 2009 & \cellcolor{green!15}82.2 &\cellcolor{green!15}- &\cellcolor{green!15}74.0 &\cellcolor{green!15}- & \cellcolor{green!15}PSMs - temporal concatenation, LDA \\
    		\hhline{*{7}{-}}
    
    		\textbf{\textit{KSVD-SHMM}} \scriptsize \cite{Tao2012, Ahmidi2017} & 2012 
    		&\cellcolor{green!15}83.4 &\cellcolor{green!15}- &\cellcolor{green!15}73.5 &\cellcolor{green!15}-  &\cellcolor{green!15}PSMs\\
    		\hhline{*{7}{-}}
    		
    		\textbf{\textit{SDSDL}} \scriptsize \cite{Sefati2015} & 2015 
    		&\cellcolor{green!15}86.3 &\cellcolor{green!15}- &\cellcolor{green!15}78.7 &\cellcolor{green!15}-  &\cellcolor{green!15}PSMs \& MTMs - mean-variance normalization, PCA \\ 
    		\hhline{*{7}{-}}
    		
    		\textbf{\textit{SDL+SC-CRF}} \scriptsize \cite{Mavroudi2018} & 2018 
    		&\cellcolor{green!15}86.2 &\cellcolor{green!15}75.5  &\cellcolor{green!15}78.5 &\cellcolor{green!15}58.0  &\cellcolor{green!15}PSMs \& MTMs - mean-variance normalization, PCA \\ 
    		\hhline{*{7}{-}}
    		
    		\multirow{3}{2.9cm}{\textbf{\textit{MsM-CRF}} \scriptsize\cite{Tao2013, Ahmidi2017}} 
    		& \multirow{3}{0.6cm}{2013}
    		& \cellcolor{green!15}82.0 &\cellcolor{green!15}- & \cellcolor{green!15}67.8 &\cellcolor{green!15}-  
    		&\cellcolor{green!15}PSMs \& MTMs - SVM scores (unary and pariwise)\\
    		& &\cellcolor{yellow!25}84.4 &\cellcolor{yellow!25}- &\cellcolor{yellow!25}77.3 &\cellcolor{yellow!25}-  
    		&\cellcolor{yellow!25} RGB frames - BoF around dense trajectories, SVM scores (unary and pairwise)\\
    		& &\cellcolor{pink!25}85.1 &\cellcolor{pink!25}- &\cellcolor{pink!25}79.1 &\cellcolor{pink!25}- &\cellcolor{pink!25} kinematics: PSMs \& MTMs - SVM scores (pairwise) \newline video: RGB frames - BoF around dense trajectories, SVM scores (unary)\\
    		\hhline{*{7}{-}}
    		
    		\textbf{\textit{STCNN+sM-CRF}} \scriptsize \cite{Lea2016} & 2016 
    		&\cellcolor{yellow!25}- &\cellcolor{yellow!25}-  &\cellcolor{yellow!25}74.2 &\cellcolor{yellow!25}66.6  &\cellcolor{yellow!25} RGB frames \& Motion History Images\\
    		\hhline{*{7}{-}}
    		
    		\multirow{2}{2.9cm}{\textbf{\textit{SC-CRF}} \scriptsize \cite{Lea2015semantic, Ahmidi2017}} 
    		& \multirow{2}{0.6cm}{2015}
    		&\cellcolor{green!15}85.2 &\cellcolor{green!15}- &\cellcolor{green!15}81.7 & \cellcolor{green!15}-  &\cellcolor{green!15} PSMs (positions, linear velocities, gripper angle)\\
    		& &\cellcolor{pink!25}85.0 &\cellcolor{pink!25}- &\cellcolor{pink!25}81.6 &\cellcolor{pink!25}-  &\cellcolor{pink!25}kinematics: PSMs (positions, linear velocities, gripper angle) \newline video: Semantic features (distance tool-object)\\
    		\hhline{*{7}{-}}
    		
    		 \textbf{\textit{LC-SC-CRF}} \scriptsize \cite{Lea2016convolutional} & 2016 &\cellcolor{pink!25}- &\cellcolor{pink!25}-  &\cellcolor{pink!25}83.5 &\cellcolor{pink!25}76.9  &\cellcolor{pink!25} kinematics: PSMs (positions, linear velocities, gripper angle) \newline  video: Semantic features (distance tool-object)\\
    		\hhline{*{7}{-}}
    		
             \textbf{\textit{CNN + LC-SC-CRF}} \scriptsize \cite{Rupprecht2016} & 2016 &\cellcolor{yellow!25}- &\cellcolor{yellow!25}-  &\cellcolor{yellow!25}76.6 &\cellcolor{yellow!25}-  &\cellcolor{yellow!25} RGB frames\\
    		\hhline{*{7}{-}}
    		
    	\end{tabular}
    	\vspace{0.2 cm}
    	
    	\begin{tablenotes}
          \small
          \item \textit{PSMs = Patient Side Manipulators, MTMs = Master Tool Manipulators, LDA = Linear Discriminative Analysis, PCA = Principal Component Analysis, SVM = Support Vector Machine, BoF = Bag of Features}
        \end{tablenotes}
        
    \end{threeparttable}

    \label{table:ResultsGraphical}
\end{table*}

\subsection{Experimental setup}

Two standard cross-validation setups have been used to analyze performance and generalization capability of the available recognition systems:

\begin{itemize}
\item \textit{Leave-one-supertrial-out (LOSO)}: five validation folds are created taking one different trial from each user. The LOSO scheme can be used to evaluate model generalization to new trials performed by known surgeons.

\item \textit{Leave-one-user-out (LOUO)}: eight validation folds are created taking all trials performed by the same user.
The LOUO scheme can be used to evaluate model generalization to new and unknown surgeons, which is more challenging due to subject-specific style variability. 

\end{itemize}

Notably, deep learning schemes typically limit extensive cross-validation due to high computational requirements and long training time and most recent studies have used only LOUO to discriminate different methods. LOUO is considered as the gold standard because it does not allow approaches to overfit to surgeon-specific features, which is also important for surgical skill assessment where networks should not learn to classify based on the surgeon but on the execution quality.

\subsection{Evaluation metrics} \label{Evaluation metrics}

Evaluation metrics for action recognition can be broadly divided into three categories highlighting different strengths and weaknesses of the proposed methods: 

\vspace{0.2 cm}
\textit{Frame-wise metrics}: standard classification metrics such as average accuracy, precision, recall and F1-score, have been widely used to assess frame-wise recognition performance. 
These were however conceived for non-sequential data and are not sufficient for an exhaustive analysis of temporal predictions and robust method comparison. 
Predictions with similar accuracy might in fact show large qualitative differences in terms of action ordering and over-segmentation (i.e. prediction of many insignificant action boundaries). 

\vspace{0.2 cm}
\textit{Segmental metrics}: 
the two most common segmental metrics, which have been specifically designed to evaluate the temporal structure of action predictions, are the \textit{edit score} and the \textit{segmental F1 score with threshold k/100} (\textit{F1@k}). 
The edit score evaluates action ordering but not their timing, penalizing out-of-order predictions and over-segmentation. It is computed as the Levenshtein distance between true and predicted label sequences, either normalized at the dataset level (Edit), where the common normalization factor is the maximum number of segments in any ground-truth sequence, or at the instance level (Edit*), where the normalization factor for each sequence is the maximum number of segments in either the ground-truth or the prediction \cite{Lea2016}.
F1@k also penalizes over-segmentation, but it additionally evaluates the temporal overlap between predictions and ground truth segments with reduced sensitivity to slight temporal shifts, compensating for annotation noise around the segment boundaries. 
F1@k is obtained by computing the intersection over union (IoU) overlap score between each predicted segment and the corresponding ground truth segment of the same class. That prediction is considered as true positive (TP) if the IoU is above a threshold $\tau = k/100$, otherwise it is a false positive (FP). TPs and FPs are then used to compute the final F1 score \cite{Lea2017tcnCVPR}.\\
Normalized mutual information (NMI) has been used less frequently to measure alignment between two action sequences in unsupervised settings \cite{Murali2016}, as this score is not affected by permutations of cluster labels. 

\textit{Unsupervised metrics}: instead of comparing predictions to their ground truth, unsupervised metrics such as the silhouette score (SS) have been used to measure the compactness of action clusters generated by unsupervised recognition methods \cite{Murali2016}. 

Other metrics used in the reviewed articles include Segmentation Accuracy (Seg-Acc) \cite{Zhao2018}, Temporal Structure Score (TSS) \cite{Goel2019} and detection scores based on temporal tolerance windows \cite{Fox2017, Tsai2019unsup}. Formulas and pseudocode of the most common evaluation metrics are reported in Appendix \ref{appendix}.

\begin{table*}[t!]
	\caption[Deep Learning.]{Average LOUO cross-validation scores (Accuracy, Edit*, Edit and F1@10) on the suturing trials of JIGSAWS - Deep Learning. green = kinematics, yellow = video, red = kinematics and video.}\label{tab2}
	\centering
	\footnotesize
	\renewcommand{\arraystretch}{1.47}
	\renewcommand{\arrayrulewidth}{1pt}
    	
    \begin{threeparttable}
    	\begin{tabular}{m{2.3cm}|P{0.6cm}|c|c|c|c|m{9.0cm}}
    		\hhline{*{7}{-}}
    		\rowcolor{white}
            \hspace{0.45cm} \textbf{Method}  &\textbf{Year} &\bfseries Acc &\bfseries Edit* &\bfseries Edit &\bfseries F1@10  &\hspace{4.3cm} \textbf{Data} \\
    		\hhline{*{7}{-}}
    		 \textbf{\textit{LSTM}} \scriptsize \cite{dipietro2016, Lea2016tcnECCV} &2016 &\cellcolor{green!15}80.5 &\cellcolor{green!15}75.3 &\cellcolor{green!15}80.2 &\cellcolor{green!15}- &\cellcolor{green!15}PSMs (positions, linear velocities, gripper angle) - 5Hz, mean-variance norm\\
    		\hhline{*{7}{-}}
    		 \textbf{\textit{BiRNN}} \scriptsize \cite{DiPietro2019}  &2019 &\cellcolor{green!15}82.1 &\cellcolor{green!15}78.9 &\cellcolor{green!15}82.7  &\cellcolor{green!15}-  &\cellcolor{green!15}PSMs (linear velocities, angular velocities, gripper angle) - 5Hz \\
    		\hhline{*{7}{-}}
    		 \textbf{\textit{BiMIST}} \scriptsize \cite{DiPietro2019}  &2019 &\cellcolor{green!15}84.7  &\cellcolor{green!15}76.4 &\cellcolor{green!15}74.3  &\cellcolor{green!15}-  &\cellcolor{green!15}PSMs (linear velocities, angular velocities, gripper angle) - 5Hz \\
    		\hhline{*{7}{-}}
    		 \textbf{\textit{BiLSTM}} \scriptsize \cite{DiPietro2019}  &2019 &\cellcolor{green!15}84.7 &\cellcolor{green!15}88.1 &\cellcolor{green!15}91.6  &\cellcolor{green!15}-  &\cellcolor{green!15}PSMs (linear velocities, angular velocities, gripper angle) - 5Hz\\
    		\hhline{*{7}{-}}
    		 \textbf{\textit{BiGRU}} \scriptsize \cite{DiPietro2019}  &2019 &\cellcolor{green!15}84.8 &\cellcolor{green!15}88.5 &\cellcolor{green!15}91.6  &\cellcolor{green!15}-  &\cellcolor{green!15}PSMs (linear velocities, angular velocities, gripper angle) - 5Hz \\
    		\hhline{*{7}{-}}
    		 \textbf{\textit{APc}} \scriptsize \cite{VanAmsterdam2020}  &2020 &\cellcolor{green!15}85.5 &\cellcolor{green!15}85.3 &\cellcolor{green!15}-  &\cellcolor{green!15}-  &\cellcolor{green!15}PSMs (positions, linear velocities, gripper angle) - 5Hz, mean-variance norm  \\
    		\hhline{*{7}{-}}
    		 \textbf{\textit{MS-RNN}} \scriptsize \cite{Gurcan2019}  &2019 &\cellcolor{green!15}90.2 &\cellcolor{green!15}- &\cellcolor{green!15}89.5 &\cellcolor{green!15}-  &\cellcolor{green!15}PSMs  \& MTMs - 5Hz \\
    		\hhline{*{7}{-}}
    		 \textbf{\textit{ST-GCN}} \scriptsize \cite{Sarikaya2020} &2020 &\cellcolor{yellow!25}67.9 &\cellcolor{yellow!25}- &\cellcolor{yellow!25}- &\cellcolor{yellow!25}-  &\cellcolor{yellow!25}RGB frames -  90-frame snippets, ResNet50 pose estimation\\
    		\hhline{*{7}{-}}
    		 \textbf{\textit{S-CNN}} \scriptsize \cite{Lea2016, Lea2016tcnECCV}  &2016 &\cellcolor{yellow!25}74.0 &\cellcolor{yellow!25}37.7 &\cellcolor{yellow!25}- &\cellcolor{yellow!25}-  &\cellcolor{yellow!25}RGB frames \& Difference Images \\
    		\hhline{*{7}{-}}
    		 \textbf{\textit{DenseNet +}} 
    		\newline \textbf{\textit{Adaptive Norm}} \scriptsize \cite{Kaku2020}  &2020 &\cellcolor{green!15}75.8 &\cellcolor{green!15}- &\cellcolor{green!15}- &\cellcolor{green!15}-  &\cellcolor{green!15}PSMs \& MTMs \\
    		\hhline{*{7}{-}}
    		 \textbf{\textit{ST-CNN}} \scriptsize \cite{Lea2016, Lea2016tcnECCV} &2016  &\cellcolor{yellow!25}77.7 &\cellcolor{yellow!25}68.0 &\cellcolor{yellow!25}- &\cellcolor{yellow!25}-  &\cellcolor{yellow!25}RGB frames \& Difference Images \\
    		\hhline{*{7}{-}}
    		\multirow{2}{2.3cm}{\textbf{\textit{TCN}} \scriptsize \cite{Lea2016tcnECCV}} 
    		&\multirow{2}{0.6cm}{2016} &\cellcolor{green!15}79.6 &\cellcolor{green!15}85.8 &\cellcolor{green!15}- &\cellcolor{green!15}-  &\cellcolor{green!15}PSMs (positions, linear velocities, gripper angle) - 10Hz\\
    		 & &\cellcolor{yellow!25}81.4 &\cellcolor{yellow!25}83.1 &\cellcolor{yellow!25}- &\cellcolor{yellow!25}- &\cellcolor{yellow!25}RGB frames \& Difference Images - 10Hz, S-CNN  \cite{Lea2016} feature extraction\\
    		\hhline{*{7}{-}}
    		 \textbf{\textit{TDNN}} \scriptsize \cite{Menegozzo2019} &2019 &\cellcolor{green!15}80.4 &\cellcolor{green!15}- &\cellcolor{green!15}- &\cellcolor{green!15}- &\cellcolor{green!15}PSMs  \& MTMs \\
    		\hhline{*{7}{-}}
    		 \multirow{2}{2.3cm}{\textbf{\textit{TCN + RL}} \scriptsize \cite{Liu2018}} 
    		&\multirow{2}{0.6cm}{2016} &\cellcolor{green!15}82.1 &\cellcolor{green!15}87.9 &\cellcolor{green!15}- &\cellcolor{green!15}91.1  &\cellcolor{green!15}PSMs (positions, linear velocities, gripper angle) \\
    		& &\cellcolor{yellow!25}81.4 &\cellcolor{yellow!25}88.0 &\cellcolor{yellow!25}- &\cellcolor{yellow!25}92.0  &\cellcolor{yellow!25}RGB frames \& Difference Images - 10Hz, S-CNN  \cite{Lea2016} feature extraction\\
    		\hhline{*{7}{-}}
    		
    		 \textbf{\textit{TCN + RL +}} 
    		\newline
    		\textbf{\textit{Tree Search}} \scriptsize \cite{Gao2020} &2020 &\cellcolor{yellow!25}81.7 &\cellcolor{yellow!25}88.5 &\cellcolor{yellow!25}- &\cellcolor{yellow!25}92.7 &\cellcolor{yellow!25} RGB frames \& Difference Images - 10Hz, S-CNN  \cite{Lea2016} feature extraction\\
    		\hhline{*{7}{-}}
    		 \textbf{\textit{MTL-VF}} \scriptsize \cite{Wang2020} &2020 &\cellcolor{yellow!25}82.1 &\cellcolor{yellow!25}86.6 &\cellcolor{yellow!25}- &\cellcolor{yellow!25}90.6  &\cellcolor{yellow!25}RGB frames - C3D \cite{Tran2015} feature extraction \\
    		\hhline{*{7}{-}}
    		 \textbf{\textit{TCED-Bd}} \scriptsize \cite{Zhang2019} &2019  &\cellcolor{yellow!25}82.2 &\cellcolor{yellow!25}87.7 &\cellcolor{yellow!25}- &\cellcolor{yellow!25}91.4  &\cellcolor{yellow!25}RGB frames \& Difference Images - 10Hz, S-CNN  \cite{Lea2016} feature extraction\\
    		\hhline{*{7}{-}}
    		 \textbf{\textit{TricorNet}} \scriptsize \cite{Ding2017} &2017  &\cellcolor{yellow!25}82.9 &\cellcolor{yellow!25}86.8  &\cellcolor{yellow!25}- &\cellcolor{yellow!25}-  &\cellcolor{yellow!25}RGB frames \& Difference Images - 10Hz, S-CNN  \cite{Lea2016} feature extraction\\
    		\hhline{*{7}{-}}
    		 \textbf{\textit{3D-CNN}} \scriptsize \cite{Funke2019} &2019  &\cellcolor{yellow!25}84.0  &\cellcolor{yellow!25}80.7 &\cellcolor{yellow!25}- &\cellcolor{yellow!25}87.2  &\cellcolor{yellow!25}RGB frames - 5Hz, 16-frame snippets, scale jittering, corner cropping \\
    		\hhline{*{7}{-}}
    		 \textbf{\textit{TDRN}} \scriptsize \cite{Lei2018} &2018  &\cellcolor{yellow!25}84.6 &\cellcolor{yellow!25}90.2 &\cellcolor{yellow!25}- &\cellcolor{yellow!25}92.9  &\cellcolor{yellow!25}RGB frames \& Difference Images - 10Hz, S-CNN  \cite{Lea2016} feature extraction\\
    		\hhline{*{7}{-}}
    		 \textbf{\textit{AT-Net}} \scriptsize \cite{Wang2019} &2019  &\cellcolor{yellow!25}- &\cellcolor{yellow!25}91.5 &\cellcolor{yellow!25}- &\cellcolor{yellow!25}94.6 &\cellcolor{yellow!25} RGB frames \& Difference Images - 10Hz, S-CNN  \cite{Lea2016} feature extraction\\
    		\hhline{*{7}{-}}
    		 \textbf{\textit{Symm dilation +}} 
    		\newline
    		\textbf{\textit{attention}} \scriptsize \cite{Zhang2020}  &2020	&\cellcolor{yellow!25}90.1 &\cellcolor{yellow!25}89.9 &\cellcolor{yellow!25}- &\cellcolor{yellow!25}92.5 &\cellcolor{yellow!25}RGB frames \& Difference Images - 10Hz, S-CNN  \cite{Lea2016} feature extraction\\
    		\hhline{*{7}{-}}
    		 \textbf{\textit{Fusion-KV}} \scriptsize \cite{Qin2020} &2020 &\cellcolor{pink!25}86.3 &\cellcolor{pink!25}87.2 &\cellcolor{pink!25}- &\cellcolor{pink!25}- &\cellcolor{pink!25}kinematics: PSMs (positions, linear velocities, Euler angles, angular velocities) \newline video: RGB frames \& Difference Images - VGG16 feature extraction\\
    		\hhline{*{7}{-}}
    		 \textbf{\textit{Fusion-KV+LSTM}} \newline
    		\textbf{\textit{(pred @1s)}} \scriptsize \scriptsize \cite{Qin2020b} &2020 &\cellcolor{pink!25}84.3 &\cellcolor{pink!25}- &\cellcolor{pink!25}- &\cellcolor{pink!25}- &\cellcolor{pink!25} kinematics: PSMs for past observation window \newline video: RGB frames for past observation window\\
    		\hhline{*{7}{-}}
		
    	\end{tabular}
        
        \vspace{0.15 cm}
    	
    	\begin{tablenotes}
          \small
          \item \textit{PSMs = Patient Side Manipulators, MTMs = Master Tool Manipulators}
        \end{tablenotes}
        
    \end{threeparttable}
    
	\label{table:ResultsDL}
\end{table*}

\subsection{Results}

Results reported on the suturing demonstrations of JIGSAWS are grouped into supervised graphical models (Table \ref{table:ResultsGraphical}), supervised deep learning (Table \ref{table:ResultsDL}), semi-supervised (Table \ref{table:ResultsSemisup}) and unsupervised (Table \ref{table:ResultsUnsup}) methods. 
Despite referring to the same dataset, attention should be paid to the input data when comparing different methods, especially when the performance gap is limited. Improvements in segmental scores, for example, could be partially attributed to the adoption of lower sampling rate rather than to the method itself. Small accuracy gaps could also be linked to specific data pre-processing (filtering, normalization, dimensionality reduction) and kinematic and visual channel selection. Given the small dataset size and similarity between certain results, more rigorous comparative analysis could be supported by statistical testing to highlight significative performance gaps. Evaluation on larger surgical datasets should however be the main goal of future researches. 

\begin{table*}[t]
	\caption[Unsupervised methods.]{Average scores on the suturing demonstrations of JIGSAWS - Unsupervised methods. \\green = kinematics, yellow = video, red = kinematics and video.\\ \ **Leave-one-out estimate.}\label{tab3}
	\centering
    \renewcommand{\arraystretch}{1.5}
    \renewcommand{\arrayrulewidth}{1pt}
	\footnotesize

    \begin{threeparttable}
    
    	\begin{tabular}{m{1.8cm}|P{0.6cm}|m{2.8cm}|m{3cm}|m{7.1cm}}
    		\hhline{*{5}{-}}
    		\rowcolor{white}
            \hspace{0.5cm} \textbf{Method} & \textbf{Year} & \hspace{0.9cm} \textbf{Scores} & \hspace{0.9cm} \textbf{Metrics} & \hspace{3.1cm} \textbf{Data} \\
    		\hhline{*{5}{-}}
    		 \textbf{\textit{TSSC-LR}} \scriptsize \cite{Clopton2017}  & 2017 & \cellcolor{green!15}49.7 / 32.8 & \cellcolor{green!15}Acc / Edit*  & \cellcolor{green!15} PSMs \& MTMs - 6 Hz, unit-norm normalization\\
    		\hhline{*{5}{-}}
    		 \multirow{3}{1.8cm}{\textbf{\textit{TSC-DL}} \scriptsize \cite{Murali2016}} & \multirow{3}{0.6cm}{2016} 
    		 & \cellcolor{green!15}0.52** / 0.31** & \cellcolor{green!15}SS / NMI  & \cellcolor{green!15} PSMs \& MTMs - 10 Hz, temporal concatenation\\
    		& & \cellcolor{yellow!25} 0.58** / 0.16** & \cellcolor{yellow!25}SS / NMI & \cellcolor{yellow!25} RGB frames - 10 Hz, cropping, rescaling, channel norm, VGG feature extraction, PCA, temporal concatenation\\
    		& & \cellcolor{pink!25} 0.73** / 0.63** & \cellcolor{pink!25}SS / NMI & \cellcolor{pink!25} kinematics: PSMs \& MTMs - 10 Hz, temporal concatenation \newline video: RGB frames - 10 Hz, cropping, rescaling, channel norm, VGG feature extraction, PCA, temporal concatenation \\
    		\hhline{*{5}{-}}
    		 \textbf{\textit{TSC-DCED-Net}} \scriptsize \cite{Zhao2018} & 2018 & \cellcolor{pink!25} 0.57** / 61.9** & \cellcolor{pink!25}NMI / Seg-Acc & \cellcolor{pink!25} kinematics: PSMs  \& MTMs - wavelet transform \newline video: RGB frames - DCED-Net feature extraction\\
    		\hhline{*{5}{-}}
    		 \textbf{\textit{DDO}} \scriptsize \cite{Fox2017} & 2017 & \cellcolor{pink!25} 72.0 & \cellcolor{pink!25} Fraction of correct boundaries (300ms tolerance) & \cellcolor{pink!25} kinematics: PSMs (positions, gripper angle) \newline video: RGB frames - rescaling\\ 
    		\hhline{*{5}{-}}
    		 \textbf{\textit{PRISM}} \scriptsize \cite{Goel2019} & 2019 & \cellcolor{green!15}46.7 / 33.3 & \cellcolor{green!15}TSS / NMI (expert only)  & \cellcolor{green!15} MTMs - 10 Hz, PCA, temporal concat, mean-variance norm \\
    		\hhline{*{5}{-}}
    		 \textbf{\textit{Soft-UGS}} \scriptsize \cite{Fard2017} & 2017  & \cellcolor{green!15}0.74 / 0.71 / 0.72 / 0.65 & \cellcolor{green!15}Recall / Precision / F1 / SS  & \cellcolor{green!15} PSMs\\
    		\hhline{*{5}{-}}
    		 \textbf{\textit{SpaceTime + \newline Variance}} \scriptsize \cite{Tsai2019unsup} & 2019 & \cellcolor{green!15} 65.2 / 92.0 / 75.4& \cellcolor{green!15}Recall / Precision / F1 & \cellcolor{green!15} PSMs (positions, orientations) - smoothing\\
    		\hhline{*{5}{-}}
    		 \textbf{\textit{Zero-shot}} \scriptsize \cite{Jones2019} & 2019 & \cellcolor{yellow!25} 56.6 / 61.7 & \cellcolor{yellow!25}Acc / Edit* & \cellcolor{yellow!25} RGB frames\\
    		\hhline{*{5}{-}}
    		
    	\end{tabular}
        \vspace{0.15 cm}
        	
        	\begin{tablenotes}
              \small
              \item \textit{PSMs = Patient Side Manipulators, MTMs = Master Tool Manipulators, PCA = Principal Component Analysis, Acc = accuracy, \\ SS = Silhouette Score, NMI = Normalized Mutual Information, Seg-Acc = Segmentation Accuracy, TSS = Temporal Structure Score}
            \end{tablenotes}
            
        \end{threeparttable}
    
	\label{table:ResultsUnsup}
\end{table*}

\begin{table*}[t]
	\caption[Semi-supervised methods.]{Average scores on the suturing demonstrations of JIGSAWS - Semi-supervised methods.\\ GT = the number of labelled trials in the training set.\\
	(*) Average over 10 runs. (**) Average over exhaustive splits.}\label{tab4}
	\centering
    \renewcommand{\arraystretch}{1.5}
    \renewcommand{\arrayrulewidth}{1pt}
	\footnotesize
    \begin{threeparttable}
    	\begin{tabular}{m{1.5cm}|P{0.6cm}|c|m{2.1cm}|m{2.7cm}|m{7.7cm}}
    		\hhline{*{6}{-}}
    		\rowcolor{white}
            \hspace{0.1cm} \textbf{Method} & \textbf{Year} & \textbf{GT} & \hspace{0.5cm} \textbf{Scores} & \hspace{0.8cm} \textbf{Metrics} & \hspace{3.4cm} \textbf{Data} \\
    		\hhline{*{6}{-}}
    		
    		\textbf{\textit{GMM0}} \scriptsize \cite{vanAmsterdam2019} & 2019 
    		& \cellcolor{green!15}3 & \cellcolor{green!15} 0.59 / 0.46 / 0.50 & \cellcolor{green!15}Acc / NMI / SS \newline (revised annotations) & \cellcolor{green!15} PSMs (positions, orientations, gripper angle, euclidean distance between PSMs) - 10 Hz, smoothing, mean-variance normalization, temporal concatenation\\
    		\hhline{*{6}{-}}
    		
    		 \multirow{2}{3cm}{\textbf{\textit{DTW + label}} \newline \textbf{\textit{transfer}} \scriptsize \cite{Gao2016align}}  & \multirow{2}{0.6cm}{2016} 
    		 & \cellcolor{green!15}5 & \cellcolor{green!15}0.67 / 0.59 / 0.69 \newline/ 0.65 (*) & \cellcolor{green!15}Acc / Precision / Recall \newline/ F1 (1s tolerance) & \cellcolor{green!15}PSMs \& MTMs - temporal concatenation, SDAE feature extraction, 
    		 temporal derivative, smoothing, downsampling\\
    		& & \cellcolor{green!15}20 & \cellcolor{green!15}0.68 / 0.63 / 0.79 \newline/ 0.70 (*) & \cellcolor{green!15}Acc / Precision / Recall \newline/ F1 (1s tolerance) & \cellcolor{green!15}PSMs \& MTMs - temporal concatenation, SDAE feature extraction, 
    		temporal derivative, smoothing, downsampling \\
    		\hhline{*{6}{-}}
    		
    		 \multirow{2}{3cm}{\textbf{\textit{LSTM}} \scriptsize \cite{DiPietro2019one}} & \multirow{2}{0.6cm}{2019} 
    		 & \cellcolor{green!15}1 & \cellcolor{green!15}70.4 / 78.2 (**) & \cellcolor{green!15}Acc / Edit  & \cellcolor{green!15} PSMs (linear velocities, angular velocities, gripper angle) - 5Hz\\
    		& & \cellcolor{green!15}7 & \cellcolor{green!15}82.4 / 89.0 (**) & \cellcolor{green!15}Acc / Edit  & \cellcolor{green!15}PSMs (linear velocities, angular velocities, gripper angle) - 5Hz\\
    		\hhline{*{6}{-}}
    		
    	\end{tabular}
        \vspace{0.15 cm}
        	
        	\begin{tablenotes}
              \small
              \item \textit{PSMs = Patient Side Manipulators, MTMs = Master Tool Manipulators, SDAE = Stacked Denoising Autoencoder, Acc = accuracy, \\NMI = Normalized Mutual Information, SS = Silhouette Score}
            \end{tablenotes}
            
        \end{threeparttable}
        
	\label{table:ResultsSemisup}
\end{table*}

\section{Discussion and open questions} \label{Discussion and open questions}

Automatic recognition of surgical gestures is a complex task and a number of considerations affect the current state of the art in available methods as detailed below.

\subsection{Robust temporal modelling}
Our comparative analysis on the JIGSAWS dataset revealed that recent deep learning models are able to capture complex temporal dependencies in surgical motion, leading to state-of-the-art recognition performance. 
In particular, hierarchical temporal convolutions and recurrent modules can model multi-scale temporal information for robust classification and boundary identification, especially when integrated with temporal attention mechanisms \cite{Zhang2020, Qin2020b}. 
Only a few studies, however, have considered the continuous nature of surgical motion and explicitly modelled gradual transitions between action segments \cite{Fard2017, Fox2017, Tsai2019, Tsai2019unsup}. Transition handling has been alternatively addressed by under-penalizing classification errors to take into account uncertainties in ground truth annotations \cite{Gao2016align}. Promising results have also been achieved with reinforcement learning, but related work in this area is still very limited and requires deeper analysis. 

As for the lookahead time, the majority of reported methods relies on a variable number of future frames for robust recognition. While real-time performance or ability to predict future samples are not always needed and pursued, further efforts in this direction could broaden the applicability of such algorithms within the surgical theatre, supporting systems for surgical process monitoring and anomaly detection and prevention \cite{Yasar2020, Qin2020b}. 

\subsection{Discriminative feature extraction} \label{Discriminative feature extraction}

Robust analysis and understanding of fine-grained surgical gestures is especially complicated by the high variability of data with same action class.
Gestures represent generic blocks of motion which can be shared across different tasks, phases and procedures, and significant variation comes from such context including the instrument types and the patient-specific environment. Other variability factors comprise the surgeon's skill level, experience and individual style, which can significantly alter the duration, kinematics and order of actions in a subject-specific manner \cite{Ahmidi2017}. For example, novice surgeons tend to make different mistakes and perform multiple gesture attempts and adjustment motions, while expert demonstrations are generally faster and smoother. Even among experts, differences might be notable in terms of surgical style, which are then reported among novice surgeons mimicking their tutors.

Similarity of data with different action class is also problematic because the amount of available information to discriminate between fine-grained gestures is limited.
Considering the suturing task, for example, the same surgical tools and objects (a needle and a thread) are operated within the same environment during the whole task execution, conferring similar visual appearance to all the constituent segments (Fig. \ref{fig:grammar}). Even the kinematic patterns of different gestures are sometimes very similar (e.g. when a surgeon attempts to push a needle through the tissue multiple times, the gestures ``positioning needle tip on insertion point" and ``pushing needle through the tissue" can be easily confused). Additional data such as tool usage information and video of the whole operating room, which can help to discriminate high-level surgical phases \cite{Twinanda2017}, are instead redundant for fine-grained analysis. 

Discriminative feature extraction from surgical video and kinematics has thus been an important research focus area. Most of the reviewed methods, however, analyze all data samples independently or consider local neighborhoods in time. 
Only few studies integrate discriminative feature extraction and high-level temporal modelling through end-to-end training \cite{Mavroudi2018}, thus embedding long-range temporal information in the low-level features. 
Nevertheless, end-to-end training of large architectures such as CNN-LSTMs, often applied to the analysis of high-level surgical workflow \cite{Yu2018, Yengera2018}, is problematic on very small datasets like JIGSAWS.

Discriminative information could also derive from optical flow motion features, which have aided surgical phase recognition \cite{Quellec2014} and action classification \cite{Simonyan2015, Sarikaya2019}, or from semantic analysis of the video data \cite{Lea2015semantic}. 
Semantic visual features could be potentially used to represent information about the surgical environment and manipulated objects that are not explicitly captured by abstract video representations or by the kinematic data, such as identification and localization of anatomical structures, measures of tissue deformation, pose or state of manipulated objects, presence of bleeding, smoke or other surgical tools operated by a surgical assistant.
In addition, semantic information could also be gained indirectly through multi-task or transfer learning with appropriate auxiliary tasks. Multi-task learning has been extensively explored at coarse granularity levels, where systems detecting surgical tools or predicting the remaining time of surgery have also shown improved phase recognition capabilities \cite{Twinanda2017, Mondal2019, Li2017}. 
Gesture recognition networks have only recently been enhanced with parallel branches for progress \cite{VanAmsterdam2020}, skill scores \cite{Wang2020} or surgical tool trajectory \cite{Qin2020b} prediction, and other auxiliary tasks could potentially be conceived to aid recognition at fine granularity (e.g. tasks based on semantic image segmentation, object tracking or anomaly detection).

\subsection{Multi-modal data integration}

The integration of multi-modal data could also play an important role in improving the performance of available recognition systems, as different sensor modalities often measure complementary information. Kinematic data, for example, describe poses and velocities of the surgical tools in the three-dimensional space, whereas visual features carry implicit complementary information about the surgical environment and instrument-tissue interactions. The combination of kinematic and visual features has in fact yielded consistent improvement over the two individual modalities \cite{Tao2013, Lea2015semantic, Murali2016, Qin2020}. Force and torque signatures have also been used for motion decomposition and skill assessment \cite{Selvaggio2018, Rosen2001, Rosen2002}, even though this information may not be readily available from system manufacturers. This lack of data access has also meant that multi-modal data integration for surgeme recognition has not been investigated thoroughly. 

Kinematic and visual information have been combined at the input level \cite{Murali2016, Zhao2018, Lea2015semantic}, at intermediate level \cite{Qin2020b} or in the final prediction \cite{Qin2020} and pruning stages \cite{Murali2016, Zhao2018}. Concatenation of multi-modal data at low processing levels might be suboptimal and inefficient, as the two sensing modalities are not directly comparable due to differences in cardinality, semantics and stochasticity \cite{Murali2016} (e.g. visual features are typically much larger than kinematic samples and more noisy). 
A splicing weight between kinematic and visual features has been used to find appropriate importance levels in different surgical tasks \cite{Zhao2018}, but it entails extensive tuning and can not adapt dynamically to environmental changes.
Class-specific weights for multi-modal late fusion \cite{Qin2020} are also unable to capture intra-gesture dynamic properties.
Improved integration could be achieved with flexible systems that seek timely multi-modal cooperation at multiple processing levels, with adaptive weighting or knowledge transfer from the strongest to the weakest modality.

\subsection{Translational research}

Despite recent advances, the field and applicability to real surgical scenarios have been greatly hindered by the lack of large and diverse datasets of annotated demonstrations, which are essential for robust training of modern recognition systems based on machine learning and deep learning. 
Current methods need testing on varied surgical tasks and procedures, as well as on data from real interventions with complex environments, blood, specularities, camera motions, illumination changes, occlusions and higher variability in motion and gesture ordering \cite{Qin2020, Luongo2020, Bawa2020}.
International collaborations for clinical data collection and sharing would not only accelerate the generation of such datasets, but also allow the representation of variations in human physiology linked to ethnicity as well as a wider spectrum of surgical techniques. 
Development of a common fine-grained surgical language with standardized decomposition criteria for different procedures and target anatomical structures is then needed in order to integrate information from multiple data sources.

It is however not trivial to define a comprehensive gesture dictionary and meaningful segmentation criteria which are consistent in the presence of unconstrained movement and generalizable to a variety of anatomies and surgical styles.
The characterization of ``surgical gesture'' is in fact an open problem itself, in particular for bimanual operations. Datasets like JIGSAWS regard fine-grained gestures as short action segments performed by either of the two arms, while the motion of the other arm is ignored. This strategy allows for faster labelling, but suffers from ambiguities on complex demonstrations where the two arms might cooperate or perform different actions at the same time. Such ambiguity can be resolved with finer-grained analysis and multi-label segmentation with separate classification of right and left arm motions. Multi-label analysis is burdensome but more precise and can additionally account for action combinations \cite{Chen2018}.

The optimal decomposition strategy could also depend on application requirements. A robotic collaborative assistant might not need to recognize highly detailed surgical motions and localize precise action boundaries in order to provide the surgeon with timely support. 
On the other side, robust understanding of task workflow, bimanual cooperation and action transitions are required to achieve surgical automation and successfully execute complex action sequences in real case scenarios.
As for surgical skill assessment, the proper decomposition strategy could depend on the annotators' evaluation technique. Identification of known skill-related gestures and failure modes while ignoring style-related variability might be beneficial.

Even when the action dictionary is well defined, however, manual labelling of surgical gestures is extremely costly, time consuming, prone to errors and inconsistencies.
To mitigate data requirements, the recognition problem can be tackled in a semi-supervised or unsupervised manner, where labels are only necessary for model testing. Even if not yet able to outperform supervised methods, these have a lot of potential to capitalise on large amounts of unlabelled data \cite{Fu2019}. 

When collection of clinical records is problematic, one of the most viable strategies to generate large-scale surgical data is represented by virtual reality simulators \cite{Bovo2017}, which can nowadays reproduce real surgical environments with impressive realism, as well as provide complementary information (e.g. surgical tool kinematics or semantic segmentation of the 3D scene) to aid recognition.
Further efforts could then be directed towards new strategies for knowledge transfer between surgical environments (simulation vs real surgery), as well as between surgical tasks \cite{Tsai2019, Rupprecht2016} or surgical robots \cite{Madapana2020}, in order to exploit cross-domain information to improve generalization performance.

Finally, an emerging and closely related research sub-field is represented by the detection and forecast of surgical errors in fine-grained gestures.
It has been recently shown that common gesture-specific errors can be identified in real-time if provided with corresponding real-time gesture labels \cite{Yasar2020}.
Public availability of error annotations \cite{Yasar2020} will hopefully stimulate future research in this relatively unexplored field, which is fundamental for several CAI implementations including workflow monitoring and surgical automation.
 
\section{Conclusion}

In this paper, we have presented a comprehensive analysis and critical review on the recognition of surgical gestures from intraoperatively collected information. Automatic recognition of surgical gestures is challenging due to the variability, complexity and fine-grained structure of surgical motion. The digitization of surgical video and increased availability of sensor information in robotic instrumentation has driven the emergence and evolution data-driven recognition models. While increasing the robustness of automatic recognition, deep-learning-based models still face many challenges compounded by the limited availability and small size of surgical training datasets that do no capture the domain variability of clinical practice across diverse and sometimes procedure specific activities. 
The field would also benefit from standardized definition of the segmentation criteria for different applications and surgical procedures to support comparative analysis and research across multiple datasets.
Technically, unsupervised and semi-supervised approaches, which do not require large amounts of ground truth labels, are important directions for future research in order to enable scaling of methods across surgical procedures and allow adaptability to the evolution of surgical technique and instrumentation.

\appendices
\section{Formulas of the most common evaluation metrics} \label{appendix}

\noindent \textbf{\textit{Accuracy (Acc)}}: Given a sequence of length N, \\
$$Acc=\frac{Nc}{N}*100, \ \ 0 < Acc < 100 $$
where $Nc$ is the number of correctly labelled frames.\\

\noindent \textbf{\textit{Edit score (Edit)}}: Given a sequence of labels $\tau$ and corresponding ground truth $\gamma$: \\
$$Edit=1-\frac{D(\tau,\gamma)}{N}*100, \ \ 0 < Edit < 100 $$
where $D$ is the Levenshtein distance between $\tau$ and $\gamma$, $N$ is either the maximum number of segments in $\tau$ or $\gamma$ (instance level normalization), or in any ground truth sequence (dataset level normalization).\\

\noindent \textbf{\textit{Segmental F1 score with threshold k/100 (F1@k)}}: Given a sequence of predicted segments $PS$ and corresponding ground truth segments $TS$: \\

\noindent \indent $TP$ = $FP$ = $0$\\
\indent $N_{used}$ = $0$\\
\indent \textbf{for} each predicted segment $PS_j$ in $PS$:\\
\indent \indent $c$ = label of $PS_j$\\
\indent \indent $TS^c$ = list of $TS$ segments of class $c$\\
\indent \indent $IoU_{all}$ = $IoU(PS_j, TS^c)$\\
\indent \indent $IoU_i$ = $max(IoU_{all})$\\
\indent \indent $TS^c_i$ = segment in $TS^c$ corresponding to $max(IoU_{all})$\\
\indent \indent \textbf{if} $IoU_i > k/100$ and $TS^c_i$ not $used$:\\
\indent \indent \indent $TP$ = $TP + 1$\\
\indent \indent \indent Flag $TS^c_i$ as $used$ \\
\indent \indent \indent $N_{used}$ = $N_{used}$ + 1 \\
\indent \indent \textbf{else}:\\
\indent \indent \indent $FP$ = $FP + 1$\\
\indent $FN$ = $length(TS)$ - $N_{used}$ \\
\indent $precision$ = $TP / (TP+FP)$\\
\indent $recall$ = $TP / (TP+FN)$\\
\indent $F1@k$ = $2 * (precision*recall) / (precision+recall) * 100$\\

\noindent where $IoU$ is the intersection over union overlap score, $TP$ are true positives, $FP$ false positives and $FN$ false negatives. Detailed implementation is publicly available \cite{LeaTcnCode}.\\

\noindent \textbf{\textit{Normalized Mutual Information (NMI)}}: Given two sequences of labels $\tau$ and $\gamma$:
$$NMI(\tau,\gamma)=\frac{I(\tau,\gamma)}{\sqrt{H(\tau)H(\gamma)}}, \ \ 0 < NMI(\tau,\gamma) < 1 $$\\
\noindent where $I$ is the mutual information and $H$ the entropy.\\

\noindent \textbf{\textit{Silhouette Score (SS)}}: The SS for a single data sample $i$ is defined as:
$$ s(i)=\frac{b(i)-a(i)}{max\{a(i),b(i)\}}, \ \ -1 < s(i) < 1 $$
\noindent where $a(i)$ is the distance (e.g. Euclidean distance) between that sample and the mean of its own cluster, while $b(i)$ is the distance between that sample and the mean of the nearest cluster it is not a part of. 
$SS$ is the average Silhouette score over all Ns samples:
$$ SS=\sum_{i=1}^{Ns}{s(i)}, \ \ -1 < SS < 1 $$
\noindent $SS$ can be normalized to $0 < SS < 1$.

\bibliography{TBMEbibliography}

\end{document}